\documentclass[10pt,twocolumn,letterpaper]{article}

\usepackage[pagenumbers]{cvpr} 

\usepackage[table]{xcolor} 
\definecolor{lightgray}{gray}{0.9} 

\definecolor{cvprblue}{rgb}{0.21,0.49,0.74}
\usepackage[pagebackref,breaklinks,colorlinks,citecolor=cvprblue]{hyperref}
\usepackage{marvosym}

\title{Harnessing Diffusion Models for Visual Perception with Meta Prompts}

\author{Qiang Wan$^1$\quad Zilong Huang$^2$\renewcommand{\thefootnote}{\Letter}\footnotemark[1]\quad Bingyi Kang$^2$\quad Jiashi Feng$^2$\quad Li Zhang$^1$\renewcommand{\thefootnote}{\Letter}\footnotemark[1]
\\
$^{1}$ Fudan University \qquad $^{2} $ ByteDance Inc
\vspace{.5em} 
\\
\textcolor{black}{\url{https://github.com/fudan-zvg/meta-prompts}}
}

\begin{document}
\maketitle
\renewcommand{\thefootnote}{\Letter}
\footnotetext[1]{Zilong Huang (zilonghuang@bytedance.com) and Li Zhang (lizhangfd@fudan.edu.cn) are the co-corresponding authors.}
\begin{abstract}
The issue of generative pretraining for vision models has persisted as a long-standing conundrum. At present, the text-to-image (T2I) diffusion model demonstrates remarkable proficiency in generating high-definition images matching textual inputs, a feat made possible through its pre-training on large-scale image-text pairs. This leads to a natural inquiry: can diffusion models be utilized to tackle visual perception tasks? 
In this paper, we propose a simple yet effective scheme to harness a diffusion model for visual perception tasks. 
Our key insight is to introduce learnable embeddings (meta prompts) to the pre-trained diffusion models to extract proper features for perception.
The effect of meta prompts are two-fold. First, as a direct replacement of the text embeddings in the T2I models, it can activate task-relevant features during feature extraction. Second, it will be used to re-arrange the extracted features to ensures that the model focuses on the most pertinent features for the task on hand.
Additionally, we design a recurrent refinement training strategy that fully leverages the property of diffusion models, thereby yielding stronger visual features. Extensive experiments across various benchmarks validate the effectiveness of our approach. Our approach achieves new performance records in depth estimation tasks on NYU depth V2 and KITTI, and in semantic segmentation task on CityScapes. Concurrently, the proposed method attains results comparable to the current state-of-the-art in semantic segmentation on ADE20K and pose estimation on COCO datasets, further exemplifying its robustness and versatility. 
\end{abstract}    
\section{Introduction}
\label{sec:intro}
\begin{figure}
  \centering
   \includegraphics[width=\hsize]{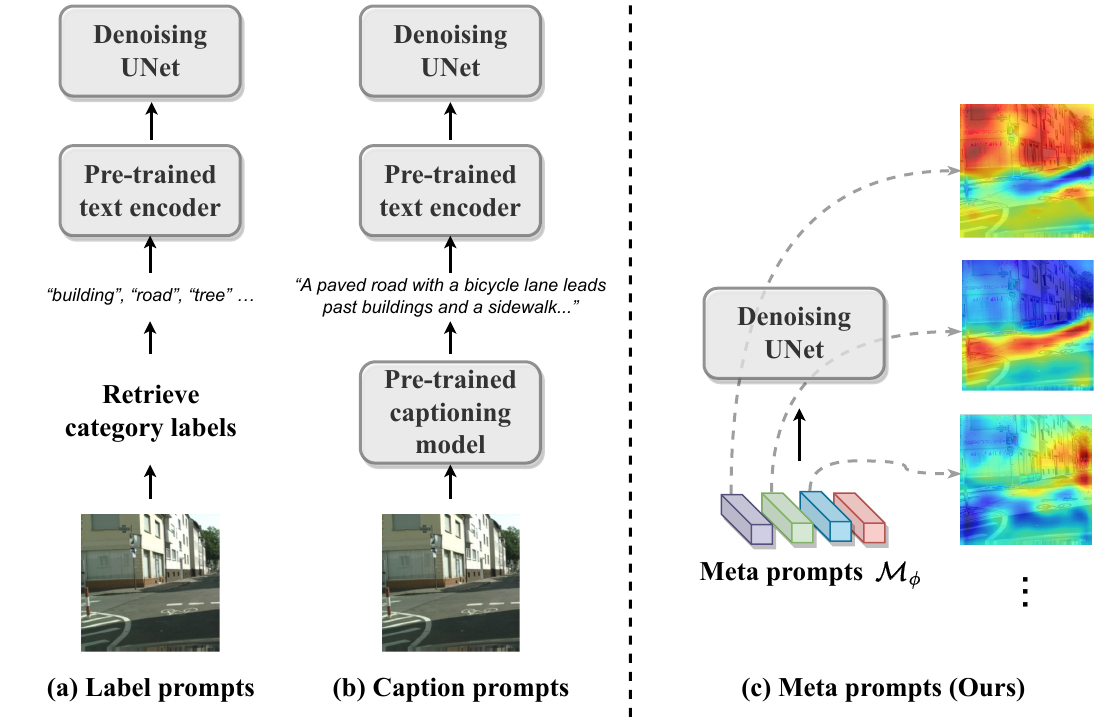}
  \caption{
  The comparison between different prompt methods. Our proposed \textbf{meta prompts} have a concise pipeline, requiring no additional labels or pre-trained models, and adaptively learn the areas of focus needed for the current perceptual task to generate corresponding activation.
  }
  \label{fig:prompt}
\end{figure}
In the continuously evolving landscape of computer vision, generating images from textual descriptions has witnessed remarkable advances, thanks to innovative models like the text-to-image diffusion model. Such models have the prowess to generate detailed images derived solely from the textual information, a feat made possible through rigorous training on voluminous image-text pair datasets. As impressive as this generative task might be, the horizon of these models' potential extends far beyond. Some pioneering works, such as~\cite{ji2023ddp, chen2023diffusiondet, duan2023diffusiondepth, lai2023denoising}, have effectively harnessed diffusion model methodologies for visual perception tasks. These innovative approaches metamorphose dense prediction tasks into a process that begins with diffusion and culminates in continuous denoising. Notably, these methods often rely on conventional backbones prevalent in visual perception tasks, extracting features from the input image to bolster the subsequent denoising process.

Yet, amidst the spectrum of methodologies, another emerging trend stands out: employing text-to-image diffusion models as feature extractors.
However, adapting the text-to-image diffusion model for visual perception tasks remains a significant challenge, particularly when the prompt interface necessitates intricate adaptation techniques.
Existing methods involve utilizing dataset category labels in textual format~\cite{zhao2023unleashing} or resorting to additional models like BLIP-2~\cite{li2023blip} to generate image captions~\cite{kondapaneni2023text}, feeding category labels or image captions into a text encoder to obtain text embeddings. 
These great works, however, may also present certain drawbacks.
Category labels as prompts fall short in benchmark evaluation due to their reliance on dataset labels, making them unsuitable for label-absent datasets. 
The limitation of utilizing image captions as prompts is that the captions produced might not align with downstream tasks. Additionally, this caption generation approach depends on an externally trained captioning model, increasing both training and inference costs.
These approaches, although effective, beckons the question: can there be an even more streamlined and efficient adaptation method?

\begin{figure*}
  \centering
   \includegraphics[width=\hsize]{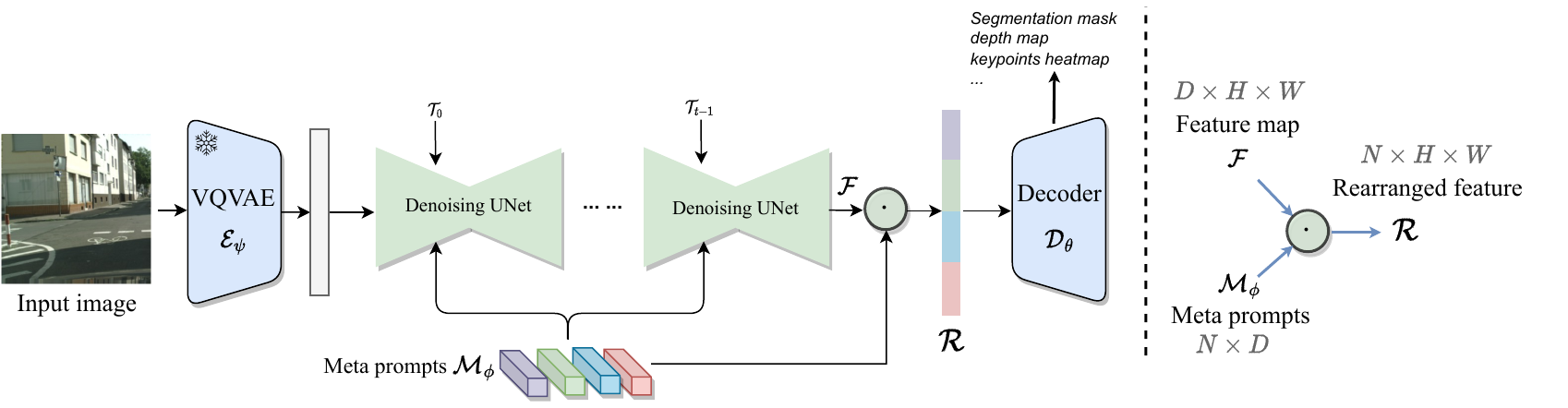}
  \caption{
  \textbf{Left: }The overall pipeline. VQVAE encoder $\mathcal{E}_\psi$ encodes the input image to latent space $Z_0$. The resolution of $Z_0$ is 1/8 of the original resolution. We refine the image feature over $t$ steps. During different refinement steps, the denoising UNet shares the same parameters. The symbol $\odot$ denotes dot production.
  \textbf{Right: }Prompts guided feature rearrangement, taking one scale feature map as an example.
  }
  \label{fig:pipeline}
\end{figure*}
In this paper, we introduce {\em perceptual diffusion model with meta prompts}, as shown in Figure~\ref{fig:pipeline}, a concise and effective take on harnessing the raw power of diffusion models for visual perception. 
Instead of relying on additional multi-modal models to generate image captions or using category labels included in datasets, we employ learnable embeddings, named as meta prompts. 
The key distinctions among different prompts interfaces are depicted in Figure~\ref{fig:prompt}.
These meta prompts can be end-to-end trained based on the target tasks and datasets, creating an adaptive condition tailored for the denoising UNet. 
They are strategically used to activate features relevant to the specific perception task within the diffusion model framework. 
Besides, the meta prompts are used to further re-arrange the extracted the hierarchical pyramid features generated by the denoising UNet by dot-production. This rearranged filtration ensures that the model focuses on the most pertinent features for the task, enhancing the accuracy and effectiveness in various visual perception applications.
After this prompts guided feature rearrangement, a specific decoder is then used to produce the final prediction results. 
Moreover, different from the manner akin to traditional perceptual model backbones that extract features with a single-forward process, we took inspiration from the diffusion model's application in generative tasks and explored the step-by-step refinement training strategy of the denoising UNet. 
As the model progresses through the refinement cycles, the distribution of the input features experiences shifts but the parameters of the UNet remain unchanged.
We address this incongruity by introducing unique timestep embeddings in each step to modulate the parameters of UNet.
Experimental results across multiple benchmarks have validated the effectiveness of our perceptual diffusion model with meta prompts. 

We make the following \textbf{contributions}:
\textbf{(i)}
We present a streamlined and impactful adaptation strategy that fine-tunes the diffusion model for visual perception tasks using a limited set of meta prompts.
\textbf{(ii)}
We have developed a recurrent refinement training approach that capitalizes on the UNet's characteristic of having identical input and output shapes, resulting in the generation of enhanced visual features.
\textbf{(iii)}
Our approach establishes new performance benchmarks in depth estimation on NYU Depth V2 and KITTI datasets, as well as in semantic segmentation on CityScapes. It also delivers results comparable to the current state-of-the-art for semantic segmentation on ADE20K and pose estimation on COCO datasets, further showcasing its robustness and versatility.

\section{Related work}

\paragraph{Diffusion model}
Diffusion models~\cite{ho2020denoising, song2020denoising, liu2022pseudo, lu2022dpm} have solidified their position as pivotal entities in image synthesis. 
Central to their architecture is the unique methodology of deliberately introducing noise into data and then iteratively restoring the original data in reverse~\cite{sohl2015deep, ho2020denoising}. 
This denoising procedure provides a novel paradigm for various applications, most prominently in generating high-quality samples. 
Their strength lies in the ability to model complex data distributions, but they are often critiqued for their computational intensity and elongated training processes. 
Building on the foundation of traditional diffusion models, latent diffusion models~\cite{rombach2022high} introduce an additional layer of complexity and flexibility by operating in a latent space. 
This configuration allows for greater manipulation of the data, often yielding improved sample quality and more precise reconstructions. 
By working in the latent space, these models can harness underlying patterns and relationships in the data more effectively. Additionally, they innovatively incorporated conditions through the cross-attention mechanism. 
Such strides have enabled the development of text-to-image diffusion models on extensive image-text pair datasets like LAION-5B~\cite{schuhmann2022laion}. 
Leveraging inherent knowledge in text-to-image diffusion models to augment visual perception tasks holds significant promise. 

\paragraph{Visual perceptual diffusion model}
Numerous innovative and impactful efforts have been advanced with the aim of translating the remarkable achievements of diffusion models in generative tasks to the complex arena of visual perception tasks.
DDP~\cite{ji2023ddp} presents a framework leveraging a conditional diffusion process for visual perception tasks, and offers dynamic, uncertainty-aware inference without the need for task-specific architectural modifications.
In~\cite{chen2023diffusiondet}, they present DiffusionDet, a framework that treats object detection as a denoising process, enhancing flexibility in box detection and iterative refinement.
DiffusionDepth~\cite{duan2023diffusiondepth} introduces an innovative method that approaches monocular depth estimation through a denoising diffusion process in latent space, refining depth predictions iteratively for greater accuracy and detail.
In~\cite{lai2023denoising}, this study enhances semantic segmentation using diffusion models, highlighting the importance of what to noise and the benefits of a simpler denoising approach for better results.
These techniques reinterpret certain perception challenges, shaping the modeling of predictive targets like segmentation masks, depth maps, and bounding boxes through diffusing and denoising processes. 
Despite adapting diffusion models to perception tasks has yielded promising results, they are notably dependent on pre-trained discriminative models from traditional perception tasks, which are utilized as image encoders to facilitate the modeling procedure.
Another emerging approach uses text-to-image diffusion models directly as image encoders. 
To better adapt text-to-image diffusion models to downstream perception tasks, these methods~\cite{zhao2023unleashing, kondapaneni2023text} utilize dataset category labels or image captions as text prompts. 
They then employ a text encoder to generate text embeddings, endowing them with the adaptative capacity to the current perception task and dataset.
Different from these creative works that rely on additional multi-modal models to generate image captions or use category labels included in datasets, we employ learnable embeddings, named as meta prompts to harness the raw power of diffusion models for visual perception. 

\section{Method}
\label{Met}
\begin{table*}[tb]
    \centering
  
    \begin{tabular}[b]{l l c r c c}
    \hline
    
    \hline
    
    \hline
    \textbf{Method}     & \textbf{Head}   & \textbf{Training data}   & \textbf{Params}       & \textbf{mIoU~(\textit{val})} & \textbf{mIoU~(\textit{test})} \\
    \hline
    \textit{non-diffusion-based} &&&&& \\
    
    SETR~\cite{zheng2021rethinking} & SETRUPHead~\cite{zheng2021rethinking} & City & 318M & 82.2 & - \\
    SegFormer~\cite{xie2021segformer} & SegFormer~\cite{xie2021segformer} & Mapi+City & 85M & 84.0& - \\
    Mask2Former~\cite{cheng2022masked} & Mask2Former~\cite{cheng2022masked} & City & -& 84.3 & -\\ 
    OneFormer~\cite{jain2023oneformer} & Transformer decoder~\cite{jain2023oneformer} & City & 372M& 84.6 & -\\  
    SeMask~\cite{jain2023semask}  & Mask2Former~\cite{cheng2022masked} & City & 222M& 85.0 & -\\  	
    OneFormer~\cite{jain2023oneformer} & Transformer decoder~\cite{jain2023oneformer} & Mapi+City & 372M& 85.8 & -\\  
    ViT-Adapte-L~\cite{chen2022vision} & Mask2Former~\cite{cheng2022masked} & Mapi+City & 571M& 85.8 & 85.2\\  	
    HRNet-OCR(MA)~\cite{tao2020hierarchical} & OCRHead~\cite{yuan2020object} & Mapi+City & - & 86.3 & 85.4\\  	
    InternImage-XL~\cite{wang2023internimage} & Upernet~\cite{xiao2018unified}  & Mapi+City& 368M & 86.4 & -\\
    HRNet-OCR(PSA)~\cite{liu2021polarized}  & OCRHead~\cite{yuan2020object} & Mapi+City & - & 86.9 & - \\  	
    InternImage-H~\cite{wang2023internimage} & Mask2Former~\cite{cheng2022masked} &Mapi+City & 1.2B & 87.0 & 86.1\\
    \hline
    \textit{diffusion-based} &&&&& \\
    DDPS~\cite{lai2023denoising} & SegFormer~\cite{xie2021segformer} & City & 123M& 82.9 & -\\
    DDP~\cite{ji2023ddp} & DeformableHead~\cite{ji2023ddp} & City & 209M& 83.9 & -\\
    \rowcolor{lightgray} \textbf{Ours} &   Upernet~\cite{xiao2018unified}    &City  & 912M    & 85.7  &  - \\  
    \rowcolor{lightgray} \textbf{Ours} &   Upernet~\cite{xiao2018unified}   &Mapi+City  & 912M    & 87.1  &  86.2\\  
    \hline
    
    \hline
    
    \hline
  \end{tabular}
\caption{Results on Cityscapes \textit{val} and \textit{test} set.  *City means CityScapes~\cite{cordts2016cityscapes} dataset, Mapi means Mapillary Vistas~\cite{neuhold2017mapillary} dataset.}
\label{table_city}
\end{table*}
\subsection{Overall architecture}
We employ the text-to-image diffusion model as a feature extractor for visual perception tasks.
As shown in Figure~\ref{fig:pipeline}, the input image is passed through VQVAE encoder $\mathcal{E}_\psi$ for image compression. This step reduces the resolution of the image to 1/8 of its original size, yielding the feature representation in latent space, $\mathbf{Z}_0$. 
It's important to note that the parameters of VQVAE encoder are fixed and do not participate in subsequent training. 
Following this, the derived $\mathbf{Z}_0$, which is retained without the addition of noise, is then fed into the UNet for feature extraction.

To better tailor the model to various tasks, the UNet simultaneously receives a modulated timestep embedding and multiple meta prompts, producing embeddings that retains a shape consistent with $\mathbf{Z}_0$. To harness more robust representational features, the initial output from UNet is recursively fed back into the UNet. During $i$-th loop, the UNet's parameters are modulated by the specific modulated timestep embedding $\mathcal{T}_i$.

Throughout the process, to enrich the feature spectrum, we engage in \textit{t} steps of recurrent refinement. This allows features from different layers within the UNet to integrate harmoniously. Ultimately, the multi-scale features generated by the UNet are dispatched to a decoder $\mathcal{D}_\theta$ specifically designed for the target visual task.

\subsection{Meta prompts}
\subsubsection{Learnable embeddings with cross attention}
The stable diffusion framework, utilizing the UNet architecture, innovatively synthesizes text-to-image by integrating textual embeddings into image features through cross attention. This integration ensures contextually and semantically accurate image generation. However, the diverse spectrum of visual perception tasks extends beyond this, as image comprehension poses different challenges, often lacking textual guidance, making text-driven approaches sometimes impractical.

In response to this challenge, our methodology adopts a more versatile strategy. Rather than solely depending on external textual prompts, we engineer an internal mechanism of learnable embeddings, termed as meta prompts, which are integrated into the diffusion model to adapt to perceptual tasks. These meta prompts, represented as a matrix $\mathcal{M}_\phi\in \mathbb{R}^{N\times D}$ where $N$ means the number of meta prompts and $D$ denotes dimension of prompts, are designed to emulate the structure of textual embeddings. A perceptual diffusion model equipped with meta prompts circumvents the necessity for external textual prompts such as dataset category labels or image captions, and obviates the need for a pre-trained text-encoder to derive the final text embeddings. Meta prompts can be trained end-to-end based on the targeted tasks and datasets, thereby establishing an adaptive condition specifically tailored for the denoising UNet. These prompts harbor the potential to encapsulate rich, task-specific semantic information.

Starting as random values, meta prompts are devoid of any meaningful information at initialization. However, through the course of training, the meta prompts undergo a transformative process. They learn, adapt, and evolve, encapsulating the subtle complexities requisite for the specific visual perception task at hand. Through iterative updates, they progress from mere noise to valuable semantic indicators. The flexibility of these meta prompts is further amplified when utilized in tandem with the cross-attention mechanism. Serving as substitutes for text embeddings, these meta prompts bridge the divide between text-to-image diffusion models and visual perception tasks. 

\subsubsection{Prompts guided feature rearrangement}
The diffusion model, by its inherent design, emphasizes the generation of multi-scale features in the denoising UNet that gradually focus on finer, low-level details as they progress towards the output layers. 
While such low-level details are not enough for tasks emphasizing texture and granularity, visual perception tasks often require an understanding that spans both low-level intricacies and high-level semantic interpretations. 
As such, the challenge lies not just in generating a rich array of features, but also in determining which combination of these multi-scale features can offer the best representation for the task at hand.

Enter the role of our meta prompts. These prompts, while being adaptive, are embedded with contextual knowledge specific to the dataset being used. 
This contextual awareness enables meta prompts to act as discerning filters, guiding the rearrangement process to sieve out the most task-relevant features from the myriad produced by the UNet.

We use dot production operation to bring together the richness of the UNet's multi-scale features and the task-adaptive nature of our meta prompts. The details are illustrated in Figure~\ref{fig:pipeline}.
Consider multi-scale features $\{F_i\}_{i=1}^{4}$, where each $F_i \in \mathbb{R}^{D \times H \times W}$. $H$ and $W$ mean height and width of the feature map. And meta prompts $\mathcal{M}_\phi \in \mathbb{R}^{N \times D}$. The rearranged features at each scale, $\{R_i\}_{i=1}^{4}$, are computed as follows:
\begin{align*}
R_i &= \mathcal{M}_\phi \cdot F_i \\
&= \mathbb{R}^{N \times D} \cdot \mathbb{R}^{D \times H \times W} \\
&= \mathbb{R}^{N \times H \times W}, \quad \text{for } i = 1, 2, 3, 4
\end{align*}

Finally, this enriched features are then introduced into task-specific decoder. 
Equipped with the task-aware features, it makes more informed decisions, leading to outputs like depth maps or segmentation masks that are detailed and semantically aligned, setting a new record in visual perception tasks.

\subsection{Recurrent refinement with modulated timestep embeddings}
In diffusion models, the iterative process of adding noise and then denoising forms the backbone of image generation. Inspired from this mechanism, we design a simple recurrent refinement for visual perception task. 
Without adding noise into the output features, we directly 
take the output features from the UNet and feed them back into the same model.
Nevertheless, there arises a challenge. 
As the model progresses through the loops, the distribution of the input features experiences shifts. 
But, the parameters of the UNet remain the same in different loops.

To address this incongruity, we introduce learnable timestep embeddings. 
Rather than being static, these embeddings are evolving during the training phase. 
For each loop, a unique timestep embedding is designed to modulate and adjust the parameters of UNet. 
This ensures that the network remains adaptive and responsive to the varying nature of input features across different steps, optimizing the feature extraction process and enhancing the model's performance in visual recognition tasks.
\section{Experiments}
\label{Exp}
\begin{table}[tb]
    \centering
  
    \begin{tabular}[b]{lccc}
    \hline
    
    \hline
    
    \hline
    \textbf{Method}    & \textbf{Iters}   & \textbf{Crop}       & \textbf{mIoU}\textbf{(ss/ms)} \\
    \hline
    \textit{non-diffusion-based} &    &     & \\
     
    Swin-L~\cite{liu2021swin} & 160K & $640^2$ & 52.1/53.5 \\
    ConvNeXt-XL~\cite{liu2022convnet} & 160K & $640^2$ & 53.6/54.0 \\
    InternImage-XL~\cite{wang2023internimage} & 160K & $640^2$ & 55.0/55.3 \\
    \hline
    \textit{diffusion-based}  &    &     & \\
    DDP~\cite{ji2023ddp}  & 160K & $512^2$ & 53.2/54.4 \\  
    VPD~\cite{zhao2023unleashing} &80K  &$512^2$ & 53.7/54.6\\
    TADP~\cite{kondapaneni2023text} &80K &$512^2$ & 54.8/55.9\\
    \rowcolor{lightgray} \textbf{Ours}   &80K  &$512^2$ & 55.7/56.7 \\  
    \hline

    \hline
    
    \hline
  \end{tabular}
\caption{Results on ADE20K \textit{val} set.}
\label{table_ade20k}
\end{table}
We evaluate our method on several perceptual tasks including semantic segmentation, depth estimation, and pose estimation.
First, we describe implementation details and compare results with state of the art.
We then conduct a series of ablation studies to validate the design of our method.
Each proposed component and important hyper-parameters are examined thoroughly.

\subsection{Experimental setup}
\label{sec:experimental_setup}
\subsubsection{Dataset} We perform semantic segmentation experiments over CityScapes~\cite{cordts2016cityscapes} and ADE20K~\cite{zhou2017scene}.
The mean of intersection over union (mIoU) is set as the evaluation metric. 
We utilize the popular benchmarks, NYU depth V2~\cite{silberman2012indoor} and KITTI~\cite{geiger2013vision} datasets, to assess our method's performance in depth estimation. Employing the standard splits, we measure our results using RMSE (root mean square error) and REL (absolute relative error) as our primary evaluation metrics.
For pose estimation on the COCO~\cite{lin2014microsoft} dataset, we adhere to the conventional splits for both training and evaluation. The Average Precision (AP) based on Object Keypoint Similarity (OKS) serves as our primary evaluation metric. Furthermore, we utilize the standard person detection results from \cite{xiao2018simple}.

\paragraph{CityScapes} stands as a pioneering resource in the domain of autonomous driving, specifically tailored for semantic segmentation tasks. Encompassing 5000 meticulously annotated high-definition images, it classifies urban scenes into 19 distinct categories.

\paragraph{ADE20K} dataset offers a vast spectrum of scenes, capturing the intricacies of both indoor and outdoor settings. With an impressive collection of 25K images, it is meticulously segmented into 150 unique categories. For a structured evaluation, the dataset is partitioned into20K/2K/3K for \textit{train}, \textit{val} and \textit{test}

\paragraph{NYU depth V2} stands as a multifaceted dataset tailored primarily for indoor depth estimation. Comprising both RGB and depth map pairs, it boasts a collection of 24K images designated for training, and an additional set of 645 images set aside for testing, altogether spanning 464 distinctive indoor scenes.

\paragraph{KITTI} renowned for its expansive coverage of driving scenarios, KITTI dataset is primarily designed to cater to the demands of monocular depth estimation. Offering stereo image sequences, KITTI dataset is often referenced in depth-related tasks. Notably, the Eigen split~\cite{eigen2014depth} within the KITTI dataset houses 23K images for training purposes, complemented by a test set of 697 images.

\paragraph{COCO} for pose estimation features extensive annotations of keypoints on human figures, making it a pivotal resource for training and evaluating human pose estimation models.  It encompasses the \textit{train} set, which has 57K training images. Evaluations are primarily reported on the \textit{val} split with 5K images and the \textit{test} split, comprising 20K images. 

\begin{table*}[ht]
\centering
\begin{tabular}{lccccccccccc}
\hline

\hline

\hline
\rowcolor{white}
\textbf{Method}  & \multicolumn{5}{c}{\textbf{NYU depth V2}} & \multicolumn{5}{c}{\textbf{KITTI Eigen split}} \\
\rowcolor{white}  & RMSE$\downarrow$ & REL$\downarrow$ & $\delta_1$ $\uparrow$ & $\delta_2$$\uparrow$ & $\delta_3$$\uparrow$ & RMSE$\downarrow$ & REL$\downarrow$ & $\delta_1$$\uparrow$ & $\delta_2$$\uparrow$ & $\delta_3$$\uparrow$\\
\hline
\textit{non-diffusion-based} &&&&&&&&&& \\
GEDepth~\cite{yang2023gedepth} & -	& - & - & - & -  & 2.044 & 0.048 & 0.976 & 0.997 & 0.999 \\
MAMo~\cite{yasarla2023mamo}  & -	& - & - & - & -  & 1.984 & 0.049 & 0.977 & 0.998 & 1.000 \\
DepthFormer~\cite{li2023depthformer}&0.339&0.096&0.921&0.989&0.998&2.143&0.052&0.975&0.997&0.999\\
PixelFormer~\cite{agarwal2023attention}&0.322&0.090&0.929&0.991&0.998&2.081&0.051&0.976&0.997&0.999\\
SwinV2-MIM~\cite{xie2023revealing}&0.287&0.083&0.949&0.994&0.999&1.966&0.050&0.977&0.998&1.000\\
ZoeDepth~\cite{bhat2023zoedepth} &0.270&0.075&0.955&0.995&0.999&-&0.057&-&-&-\\
MeSa~\cite{khan2023mesa} &0.238&0.066&0.964&0.995&0.999&-&-&-&-&-\\
\hline
\textit{diffusion-based} &&&&&&&&&& \\
DDP~\cite{ji2023ddp} &0.329&0.094&0.921&0.990&0.998&2.072&0.050&0.975&0.997&0.999\\
DepthGen~\cite{saxena2023monocular}&0.314&0.074&0.946&0.987&0.996&2.985&0.064&0.953&0.991&0.998\\
VPD~\cite{zhao2023unleashing}  & 0.254	& 0.069 & 0.964 & 0.995 & 0.999  & - & - & - & - & - \\
TADP~\cite{kondapaneni2023text}  & 0.225	& 0.062 & 0.976 & 0.997 & 0.999  & - & - & - & - & - \\
\rowcolor{lightgray} \textbf{Ours} & \textbf{0.223}  & \textbf{0.061} & \textbf{0.976} & \textbf{0.997} & \textbf{0.999} & \textbf{1.929} & \textbf{0.047} & \textbf{0.982} & \textbf{0.998} & \textbf{1.000} \\     
\hline

\hline

\hline
\end{tabular}
\caption{Results of depth estimation on NYU depth V2 and KITTI Eigen split. RMSE means root mean square error and REL means absolute relative error and accuracy metrics ($\delta_i < 1.25^i$ for i $\in$ 1, 2, 3)}
\label{tab:depth}
\end{table*}

\subsubsection{Implementation details}
We construct our model's backbone using the pre-trained text-to-image diffusion model, specifically leveraging the 1-5 version from the renowned "Stable-Diffusion" release. The VQVAE encoder's parameters remain unaltered during the training phase. 
For semantic segmentation, the synchronized BatchNorm is used in the segmentation head.
Our implementation is based on public codebase \texttt{mmsegmentation}~\cite{contributors2020mmsegmentation}.

\noindent{\bf Cityscapes} Our initial learning rate is set at 0.00008, paired with a weight decay of 0.001. All training images undergo random scaling before being cropped to a consistent size of 1024 × 1024. Aligned with prevalent methodologies~\cite{wang2023internimage, chen2022vision, xie2021segformer, tao2020hierarchical}, our training regimen begins with pre-training on Mapillary Vistas~\cite{neuhold2017mapillary}, followed by a fine-tuning phase on Cityscapes, which persists for 80k iterations.

\noindent{\bf ADE20K} We maintain an initial learning rate of 0.00008 and a weight decay of 0.001. A "poly" learning rate strategy with a factor of 1.0 is employed. Our batch size, training iteration scheduler, and data augmentation protocol are in line with~\cite{zhao2023unleashing,kondapaneni2023text}, ensuring a balanced comparison.

\noindent{\bf NYU depth V2} During the training process, we employ random cropping on images to attain a size of 480×480. The learning rate is 5e-4. We conduct training over 25 epochs, with each epoch accommodating a batch size of 24. Test-time procedures include the application of flips and sliding windows. The depth's maximum range is capped at 10m, and the model's head integrates multiple deconvolutions.

\noindent{\bf KITTI} Image cropping during training results in a consistent size of 352×352. Adopting a learning rate of 5e-4, we train our model across 25 epochs, with each epoch having a batch size of 24. For testing, we introduce both flip and sliding window techniques. Depth measurements are confined to a maximum of 80m.

\begin{table}[tb]
    \centering
  
    \begin{tabular}[b]{l|c|c}
    \hline
    
    \hline
    
    \hline
    \textbf{Method}   & \textbf{Backbone}     & \textbf{AP / AR} \\
    \hline
    SimpleBaseline~\cite{xiao2018simple}& ResNet152~\cite{he2016deep} &  73.5 / 79.0 \\
    HRNet~\cite{sun2019deep} & HRNet-W32 &  75.8 / 81.0 \\  
    HRNet~\cite{sun2019deep} & HRNet-W48  & 76.3 / 81.2 \\  
    UDP~\cite{huang2020devil} & HRNet-W48  &  77.2 / 82.0 \\
   HRFormer-B~\cite{yuan2021hrformer} & HRFormer-B &  77.2 / 82.0 \\
   ViTPose-L~\cite{xu2022vitpose} & ViT-L & 78.3 / 83.5 \\ 
    
    \rowcolor{lightgray} \textbf{Ours}  &SD1-5    & 79.0 / 84.0 \\  
    \hline
    
    \hline
    
    \hline
  \end{tabular}
\caption{Results on MS-COCO \textit{val} set.}
\label{table_coco}
\end{table}
\noindent{\bf COCO} Adhering to the conventional top-down paradigm for human pose estimation, a detection model identifies individual human instances, post which our model assesses the keypoints of these identified instances. For performance evaluations on the MS COCO Keypoint \textit{val} set, we rely on detection outcomes from SimpleBaseline~\cite{xiao2018simple}. Our input resolution is fixed at 576 × 448, and we employ the AdamW~\cite{reddi2019convergence} optimization strategy with a learning rate of 5e-4. Post-processing employs Udp~\cite{huang2020devil}. Training spans 210 epochs, with the learning rate undergoing a ten-fold reduction at the 170th and 200th epochs.

\subsection{Comparison with state of the art}

\paragraph{Semantic segmentation on Cityscapes}
In our comprehensive benchmarking on the Cityscapes \textit{val} and  \textit{test} set, the performance of our approach was evaluated against prevailing models, using both Parameters (Params) and Mean Intersection-over-Union (mIoU) as metrics. The results, detailed in Table~\ref{table_city}, underline the superiority of our method. When compared to the robust perceptual baseline, InternImage-H~\cite{wang2023internimage}, we achieves higher segmentation accuracy with fewer parameters. Moreover, it surpasses the ViT-based model, ViT-Adapter~\cite{chen2022vision}, by a significant + 1.3\% mIoU margin. These findings solidify the status of our model as a potent feature extractor.

\begin{table*}[t]

\centering
\subfloat[
\textbf{Different prompt method}. Meta prompts
\label{tab:adapt_method} are effective.
]
{
\begin{minipage}{0.34\linewidth}
{\begin{center}
\begin{tabular}{c|c|c}
\textbf{Prompt interface} & \textbf{mIoU} & \textbf{RMSE}~$\downarrow$ \\
\hline
Label prompts & 83.6 & 0.258 \\
Caption prompts& 83.4 & 0.225 \\
\rowcolor{lightgray} Meta prompts & \textbf{84.7} & \textbf{0.223} \\
\end{tabular}
\end{center}}
\end{minipage}
}
\hspace{0.5em}
\subfloat[
\textbf{The number of meta prompts}. 50 promts are optimal in depth estimation but 150 is best choice in semantic segmentation.
\label{tab:query_num}
]{
\begin{minipage}{0.29\linewidth}{\begin{center}
\begin{tabular}{c|c|c}
\textbf{\#Prompts} & \textbf{mIoU} & \textbf{RMSE}~$\downarrow$ \\
\hline
50 & 84.0 & \textbf{0.223} \\
100 & 84.5 & 0.226 \\
150 & {\textbf{84.7}} & 0.225 \\
\end{tabular}
\end{center}}\end{minipage}
}
\hspace{0.5em}
\subfloat[
\textbf{Refinement step}. With an increase in the number of steps, there is a corresponding rise in accuracy.
\label{tab:refine_step}
]{
\begin{minipage}{0.29\linewidth}{\begin{center}
\begin{tabular}{c|c|c}
\textbf{Refinement step} & \textbf{mIoU} & \textbf{RMSE}~$\downarrow$ \\
\hline
1 & 84.1 & 0.243 \\
2 & 84.5 & 0.227 \\
\rowcolor{lightgray} 3 & \textbf{84.7} & {\textbf{0.223}} \\
\end{tabular}
\end{center}}\end{minipage}
}
\hspace{2em}
\\
\subfloat[
\textbf{Prompts guided feature rearrangement}. Incorporating prompts guided feature rearrangement can boost performance.
\label{tab:prompt_guide}
]{
\centering
\begin{minipage}{0.37\linewidth}{\begin{center}
\begin{tabular}{c|c|c}
\textbf{Feature rearrangement} & \textbf{mIoU} & \textbf{RMSE}~$\downarrow$ \\
\hline
$\times$ & 84.2 & 0.226 \\
\rowcolor{lightgray} \checkmark & \textbf{84.7} & \textbf{0.223} \\
\end{tabular}
\end{center}}\end{minipage}
}
\hspace{3em}
\subfloat[
\textbf{The modulated timestep embeddings}. Modulated timestep embeddings achieve performance gains.
\label{tab:modulate_time}
]{
\begin{minipage}{0.4\linewidth}{\begin{center}
\begin{tabular}{c|c|c}
\textbf{Modulated timestep embeddings} & \textbf{mIoU} & \textbf{RMSE}~$\downarrow$ \\
\hline
$\times$ & 84.3 & 0.227 \\
\rowcolor{lightgray} \checkmark & \textbf{84.7} & \textbf{0.223} \\
\end{tabular}
\end{center}}\end{minipage}
}
\hspace{21em}
\caption{\textbf{Ablation experiments} on CityScapes semantic segmentation datasets and NYU depth V2 depth estimation dataset. We report mean of intersection over union (mIoU) and root mean square error (RMSE). Default settings are marked in \colorbox{lightgray}{gray}.}
\label{tab:ablations} 
\end{table*}

\paragraph{Semantic segmentation on ADE20K}
We compare with the previous approaches on ADE20K validation set in Table~\ref{table_ade20k}. From the results, it can be seen that our method is +2.0\% mIoU higher (55.7\% \textit{vs.} 53.7\%) than diffusion-based model VPD~\cite{zhao2023unleashing} and 0.9\% mIoU higher (55.7\% \textit{vs.} 54.8\%) than TADP~\cite{kondapaneni2023text}. 

\paragraph{Depth estimation on NYU depth V2}
In Table~\ref{tab:depth}, we juxtapose the performance against previous methods on the NYU Depth V2 dataset. The results reveal that our model surpasses the diffusion-based model, VPD~\cite{zhao2023unleashing} and other visual pre-traing methods, by achieving a notably lower RMSE. Specifically, using the default training schedule, we register an RMSE of 0.223, setting a new benchmark in the state-of-the-art for depth estimation on this dataset.

\paragraph{Depth estimation on KITTI}
In Table~\ref{tab:depth}, we present a comparison between prior methods and ours on the KITTI dataset. Our findings indicate that our proposed approach outperforms previous methods, by delivering a significantly reduced REL and RMSE. With the implementation of our standard training protocol, we achieve an RMSE of 1.929 and REL of 0.047, thereby establishing a fresh benchmark for depth estimation on this dataset.

\paragraph{Pose estimation on COCO}
In our detailed evaluation of perceptual diffusion model with meta prompts for pose estimation, as shown in Table~\ref{table_coco}, we rigorously compared its performance against existing methods using Average Precision (AP) and Average Recall (AR) as the primary evaluation metrics on COCO dataset.
The results from the table are quite enlightening. Our model demonstrated superior performance, as evidenced by higher AP and AR scores compared to baseline models. This improvement demonstrates the strong generalization ability of our model, showcasing robust and versatile performance across various visual tasks.
One of the key factors contributing to this enhanced performance was our model's unique approach to feature rearrangement and refinement, facilitated by the meta prompts. This approach allowed the model to focus on the most relevant features for pose estimation, adapting dynamically to the complexities of human anatomy and movement.

\subsection{Ablation study}
In this section, we ablate our method and some important design elements, including our meta prompts, prompts guided feature rearrangement, the number of meta prompts, refinement steps, and modulated timestep embeddings on NYU depth V2 dataset and CityScapes dataset.

\paragraph{Comparison with different prompts method}
In this section, to validate the effectiveness of our proposed meta prompts, we conducted experiments on depth estimation and semantic segmentation. 
We compared three promtpts methods of adapting diffusion models to perception tasks, including category label prompts, image caption prompts, and our proposed meta prompts. 
As demonstrated in the Table~\ref{tab:adapt_method}, we found that our meta prompts adaptation method, which doesn't require additional prior knowledge like dataset category labels or image captions, achieves superior results. 
The experimental outcomes attest to the effectiveness and simplicity of our approach.

\paragraph{Effectiveness of prompts guided feature rearrangement}
To highlight the importance of our prompts guided feature rearrangement design, we carried out empirical validation on depth estimation and semantic segmentation. 
The results, detailed in the Table~\ref{tab:prompt_guide}, reveal that incorporating prompts guided feature rearrangement significantly improves performance: there's a 0.5 increase in mIoU for semantic segmentation on CityScapes dataset and a 0.003 decrease in RMSE for depth estimation on NYU depth V2 dataset. 
These findings confirm the efficacy of our method.
Our proposed prompts guided feature rearrangement brings together the richness of the UNet's multi-scale features and the task-adaptive nature of our meta prompts. 

\begin{figure}[ht]
    \centering
    \begin{subfigure}[b]{0.11\textwidth}
        \includegraphics[width=\textwidth]{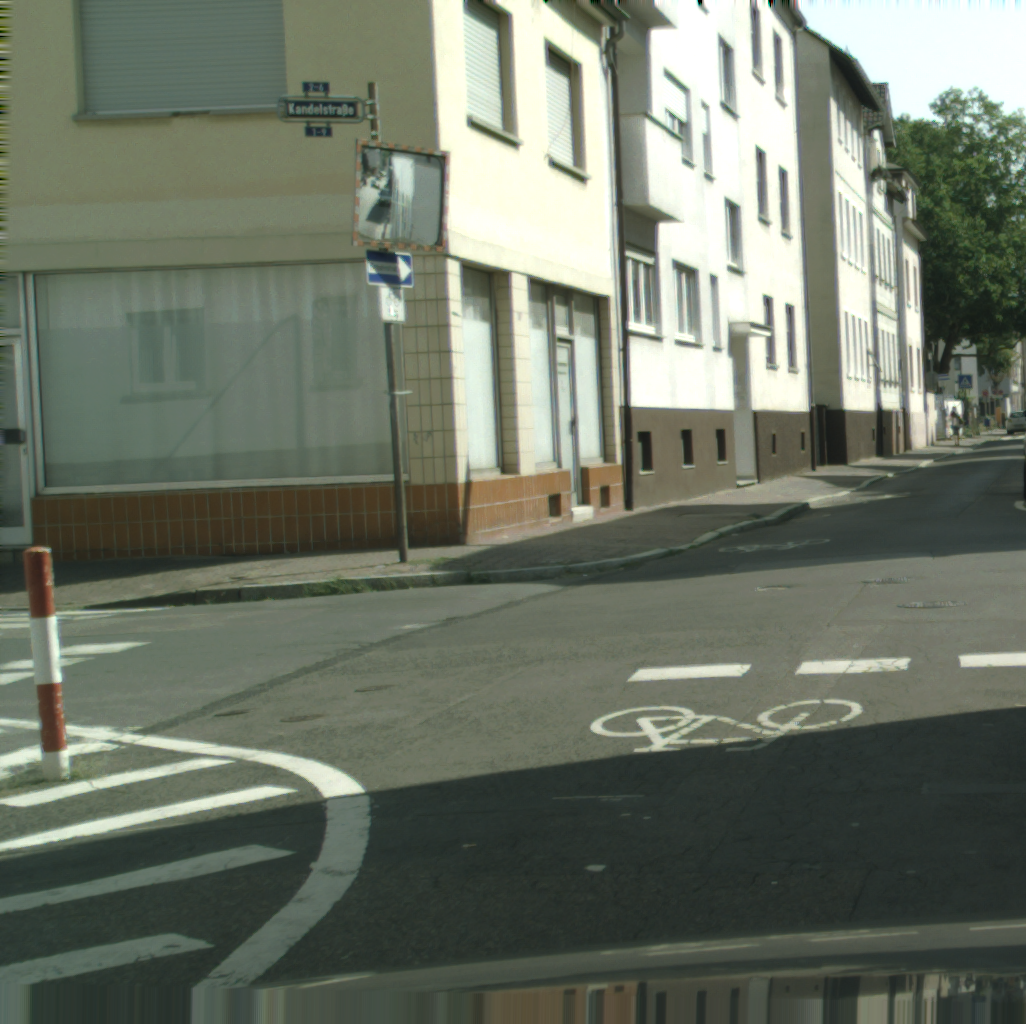}
        \caption{Input image}
        \label{fig:input_image}
    \end{subfigure}
    \begin{subfigure}[b]{0.11\textwidth}
        \includegraphics[width=\textwidth]{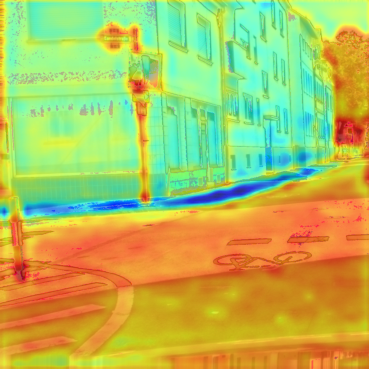}
        \caption{7th prompt}
        \label{fig:heatmap_7}
    \end{subfigure}
    \begin{subfigure}[b]{0.11\textwidth}
        \includegraphics[width=\textwidth]{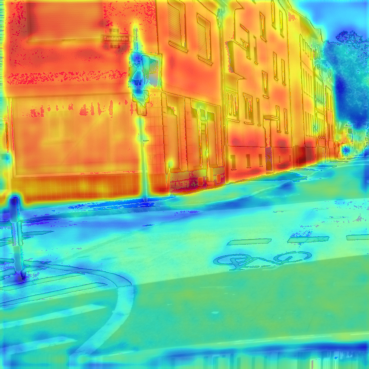}
        \caption{13th prompt}
        \label{fig:heatmap_13}
    \end{subfigure}
    \begin{subfigure}[b]{0.11\textwidth}
        \includegraphics[width=\textwidth]{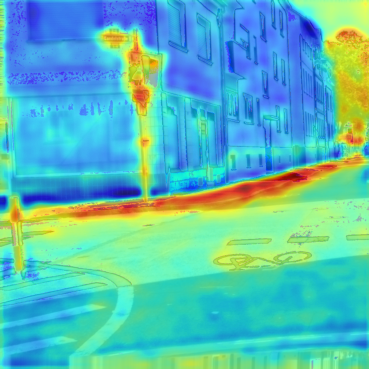}
        \caption{66th prompt}
        \label{fig:heatmap_66}
    \end{subfigure}

    \begin{subfigure}[b]{0.11\textwidth}
        \includegraphics[width=\textwidth]{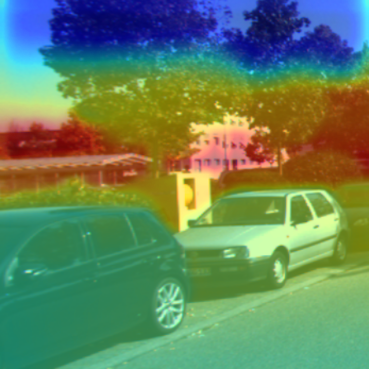}
        \caption{1th image}
        \label{fig:img1_query43}
    \end{subfigure}
    \begin{subfigure}[b]{0.11\textwidth}
        \includegraphics[width=\textwidth]{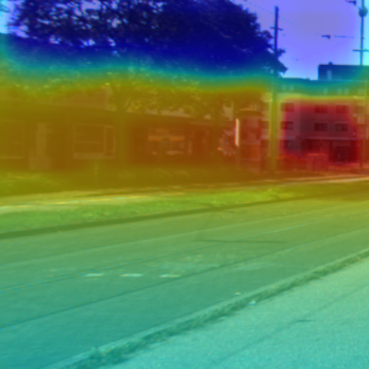}
        \caption{2th image}
        \label{fig:img2_query43}
    \end{subfigure}
    \begin{subfigure}[b]{0.11\textwidth}
        \includegraphics[width=\textwidth]{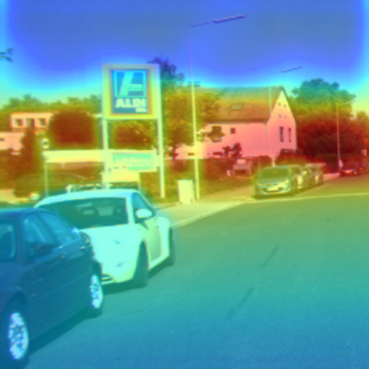}
        \caption{3th image}
        \label{fig:img3_query43}
    \end{subfigure}
    \begin{subfigure}[b]{0.11\textwidth}
        \includegraphics[width=\textwidth]{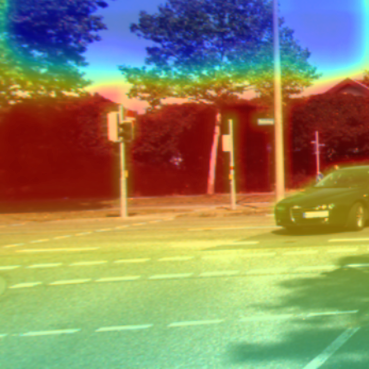}
        \caption{4th image}
        \label{fig:img4_query43}
    \end{subfigure}
    
    \caption{\textbf{First row: }Heatmaps of different prompts on the same image for semantic segmentation task. \textbf{Second row: }Heatmaps of the same prompt on different images for depth estimation task.}
    \label{fig:heatmaps}
    \vspace{-3mm}
\end{figure}

\paragraph{The influence of refinement steps and meta prompts numbers}
To study the influence of the refinement step of denoising UNet, we perform experiments with different steps on depth estimation and semantic segmentation tasks.
The results, as shown in the Table~\ref{tab:refine_step}, indicate that with the increase in the number of refinement steps, there is a corresponding enhancement in performance for both depth estimation and semantic segmentation. 
In image synthesis, as we increase the number of denoising steps, we witness a progressive enhancement in image fidelity. 
We have observed a similar effect in perception tasks, where using shared UNet parameters and increasing the number of refinement steps leads to a marked enhancement in performance.
Each iteration or cycle enables the model to delve deeper, capturing more intricate patterns and details. 

Table~\ref{tab:query_num} showcases how different quantities of meta prompts affect the model's performance. It indicates that for semantic segmentation task, 150 meta prompts is the optimal choice, while for depth estimation tasks, just 50 meta prompts are adequate.

\paragraph{The influence of modulated timestep embeddings}
As the model progresses through refinement, the distribution of the input features shifts.
Due to the fixed parameters of UNet, there arises a mismatch between the feature distributions of different steps and the consistent model parameters.
To address this incongruity, we propose modulated timestep embeddings.  
As shown in Table~\ref{tab:modulate_time}, with modulated timestep embeddings, We achieved performance gains in both depth estimation and semantic segmentation tasks.
At each step or cycle, the specific timestep embedding modulate and finetune the UNet's parameters, maintaining the adaptability of network to the evolving input feature distribution, thereby refining the feature extraction process and improving the model's capacity in visual recognition tasks.
\section{Qualitative results}

To showcase the efficacy of our method, we undertook a detailed visualization of the heatmaps generated from meta prompts. 
Specifically, these prompts are used to guide the feature rearrangement process. 
The rearranged features, once identified, are upscaled to match the original image resolution through bilinear interpolation. 
This upscaled, normalized features are then superimposed onto the original image, allowing us to create heatmaps that visually represent the areas of focus and interest.

Given the extensive number of meta prompts, reaching up to 150 in tasks like semantic segmentation, we opted for a more manageable approach in our visualization efforts. 
We focused on a representative subset, showcasing three of meta prompts for clarity and simplicity. 

In practical image inference scenarios, these prompts play a pivotal role. 
They empower the model to focus on particular facets of information that are critical for accurate predictions. 
As shown in the first row of Figure~\ref{fig:heatmaps}, for semantic segmentation task, different meta prompts may focus on various objects for classification purpose. For instance, certain prompts, like~\ref{fig:heatmap_13}, might be fine-tuned to hone in on the building or wall.
Others might skew towards identifying and understanding semantic information of road, tree or sky. 
For depth estimation task, meta prompts may exhibit some depth-aware properties. As depicted in the second row of Figure~\ref{fig:heatmaps}, when the same prompt is applied to different images, it tends to focus more on objects that are at a distance.

Each prompt operates within its own domain of expertise, yet in concert, they form a complementary and cohesive whole, significantly bolstering the model's ability to adeptly adapt diffusion models for perceptual tasks. The collective contribution of these meta prompts, as depicted in our visualizations, underscores their indispensable role in enhancing the model's interpretative and predictive capabilities in complex visual perception scenarios.
\section{Conclusion}
In this study, we introduced a straightforward yet potent adaptation method, integrating the diffusion model with a set of learnable embeddings, termed meta prompts. These prompts are adept at activating and filtering task-relevant features within the model, enhancing its utility as a feature extractor. Coupled with this, recurrent refinement training strategy, capitalizing on the UNet's consistent input-output shape, significantly strengthens the visual features produced.
Our method sets new performance benchmarks in depth estimation on datasets like NYU Depth V2 and KITTI, and in semantic segmentation on CityScapes. It also competes strongly with current state-of-the-art methods in semantic segmentation on the ADE20K dataset and pose estimation on the COCO dataset. These results not only underscore the effectiveness of our adaptation scheme but also demonstrate its robustness and adaptability.

{
    \small
    \bibliographystyle{ieeenat_fullname}
    \bibliography{main}

\begin{thebibliography}{52}
\providecommand{\natexlab}[1]{#1}
\providecommand{\url}[1]{\texttt{#1}}
\expandafter\ifx\csname urlstyle\endcsname\relax
  \providecommand{\doi}[1]{doi: #1}\else
  \providecommand{\doi}{doi: \begingroup \urlstyle{rm}\Url}\fi

\bibitem[Agarwal and Arora(2023)]{agarwal2023attention}
Ashutosh Agarwal and Chetan Arora.
\newblock Attention attention everywhere: Monocular depth prediction with skip attention.
\newblock In \emph{WACV}, 2023.

\bibitem[Bhat et~al.(2023)Bhat, Birkl, Wofk, Wonka, and M{\"u}ller]{bhat2023zoedepth}
Shariq~Farooq Bhat, Reiner Birkl, Diana Wofk, Peter Wonka, and Matthias M{\"u}ller.
\newblock Zoedepth: Zero-shot transfer by combining relative and metric depth.
\newblock \emph{arXiv preprint}, 2023.

\bibitem[Chen et~al.(2023)Chen, Sun, Song, and Luo]{chen2023diffusiondet}
Shoufa Chen, Peize Sun, Yibing Song, and Ping Luo.
\newblock Diffusiondet: Diffusion model for object detection.
\newblock In \emph{ICCV}, 2023.

\bibitem[Chen et~al.(2022)Chen, Duan, Wang, He, Lu, Dai, and Qiao]{chen2022vision}
Zhe Chen, Yuchen Duan, Wenhai Wang, Junjun He, Tong Lu, Jifeng Dai, and Yu Qiao.
\newblock Vision transformer adapter for dense predictions.
\newblock \emph{arXiv preprint}, 2022.

\bibitem[Cheng et~al.(2022)Cheng, Misra, Schwing, Kirillov, and Girdhar]{cheng2022masked}
Bowen Cheng, Ishan Misra, Alexander~G Schwing, Alexander Kirillov, and Rohit Girdhar.
\newblock Masked-attention mask transformer for universal image segmentation.
\newblock In \emph{CVPR}, 2022.

\bibitem[Contributors(2020{\natexlab{a}})]{contributors2020mmsegmentation}
MMSegmentation Contributors.
\newblock Mmsegmentation: Openmmlab semantic segmentation toolbox and benchmark, 2020{\natexlab{a}}.

\bibitem[Contributors(2020{\natexlab{b}})]{mmpose2020}
MMPose Contributors.
\newblock Openmmlab pose estimation toolbox and benchmark.
\newblock \url{https://github.com/open-mmlab/mmpose}, 2020{\natexlab{b}}.

\bibitem[Cordts et~al.(2016)Cordts, Omran, Ramos, Rehfeld, Enzweiler, Benenson, Franke, Roth, and Schiele]{cordts2016cityscapes}
Marius Cordts, Mohamed Omran, Sebastian Ramos, Timo Rehfeld, Markus Enzweiler, Rodrigo Benenson, Uwe Franke, Stefan Roth, and Bernt Schiele.
\newblock The cityscapes dataset for semantic urban scene understanding.
\newblock In \emph{CVPR}, 2016.

\bibitem[Duan et~al.(2023)Duan, Guo, and Zhu]{duan2023diffusiondepth}
Yiqun Duan, Xianda Guo, and Zheng Zhu.
\newblock Diffusiondepth: Diffusion denoising approach for monocular depth estimation.
\newblock \emph{arXiv preprint}, 2023.

\bibitem[Eigen et~al.(2014)Eigen, Puhrsch, and Fergus]{eigen2014depth}
David Eigen, Christian Puhrsch, and Rob Fergus.
\newblock Depth map prediction from a single image using a multi-scale deep network.
\newblock In \emph{NeurIPS}, 2014.

\bibitem[Geiger et~al.(2013)Geiger, Lenz, Stiller, and Urtasun]{geiger2013vision}
Andreas Geiger, Philip Lenz, Christoph Stiller, and Raquel Urtasun.
\newblock Vision meets robotics: The kitti dataset.
\newblock \emph{IJRR}, 2013.

\bibitem[He et~al.(2016)He, Zhang, Ren, and Sun]{he2016deep}
Kaiming He, Xiangyu Zhang, Shaoqing Ren, and Jian Sun.
\newblock Deep residual learning for image recognition.
\newblock In \emph{CVPR}, 2016.

\bibitem[Ho et~al.(2020)Ho, Jain, and Abbeel]{ho2020denoising}
Jonathan Ho, Ajay Jain, and Pieter Abbeel.
\newblock Denoising diffusion probabilistic models.
\newblock In \emph{NeurIPS}, 2020.

\bibitem[Huang et~al.(2020)Huang, Zhu, Guo, and Huang]{huang2020devil}
Junjie Huang, Zheng Zhu, Feng Guo, and Guan Huang.
\newblock The devil is in the details: Delving into unbiased data processing for human pose estimation.
\newblock In \emph{CVPR}, 2020.

\bibitem[Jain et~al.(2023{\natexlab{a}})Jain, Li, Chiu, Hassani, Orlov, and Shi]{jain2023oneformer}
Jitesh Jain, Jiachen Li, Mang~Tik Chiu, Ali Hassani, Nikita Orlov, and Humphrey Shi.
\newblock Oneformer: One transformer to rule universal image segmentation.
\newblock In \emph{CVPR}, 2023{\natexlab{a}}.

\bibitem[Jain et~al.(2023{\natexlab{b}})Jain, Singh, Orlov, Huang, Li, Walton, and Shi]{jain2023semask}
Jitesh Jain, Anukriti Singh, Nikita Orlov, Zilong Huang, Jiachen Li, Steven Walton, and Humphrey Shi.
\newblock Semask: Semantically masked transformers for semantic segmentation.
\newblock In \emph{ICCV}, 2023{\natexlab{b}}.

\bibitem[Ji et~al.(2023)Ji, Chen, Xie, Hong, Liu, Liu, Lu, Li, and Luo]{ji2023ddp}
Yuanfeng Ji, Zhe Chen, Enze Xie, Lanqing Hong, Xihui Liu, Zhaoqiang Liu, Tong Lu, Zhenguo Li, and Ping Luo.
\newblock Ddp: Diffusion model for dense visual prediction.
\newblock \emph{arXiv preprint}, 2023.

\bibitem[Khan et~al.(2023)Khan, Liang, Wang, Yang, and Lou]{khan2023mesa}
Muhammad~Osama Khan, Junbang Liang, Chun-Kai Wang, Shan Yang, and Yu Lou.
\newblock Mesa: Masked, geometric, and supervised pre-training for monocular depth estimation.
\newblock \emph{arXiv preprint}, 2023.

\bibitem[Kondapaneni et~al.(2023)Kondapaneni, Marks, Knott, Guimar{\~a}es, and Perona]{kondapaneni2023text}
Neehar Kondapaneni, Markus Marks, Manuel Knott, Rog{\'e}rio Guimar{\~a}es, and Pietro Perona.
\newblock Text-image alignment for diffusion-based perception.
\newblock \emph{arXiv preprint}, 2023.

\bibitem[Lai et~al.(2023)Lai, Duan, Dai, Li, Fu, Li, Qiao, and Wang]{lai2023denoising}
Zeqiang Lai, Yuchen Duan, Jifeng Dai, Ziheng Li, Ying Fu, Hongsheng Li, Yu Qiao, and Wenhai Wang.
\newblock Denoising diffusion semantic segmentation with mask prior modeling.
\newblock \emph{arXiv preprint}, 2023.

\bibitem[Li et~al.(2023{\natexlab{a}})Li, Li, Savarese, and Hoi]{li2023blip}
Junnan Li, Dongxu Li, Silvio Savarese, and Steven Hoi.
\newblock Blip-2: Bootstrapping language-image pre-training with frozen image encoders and large language models.
\newblock \emph{arXiv preprint}, 2023{\natexlab{a}}.

\bibitem[Li et~al.(2023{\natexlab{b}})Li, Chen, Liu, and Jiang]{li2023depthformer}
Zhenyu Li, Zehui Chen, Xianming Liu, and Junjun Jiang.
\newblock Depthformer: Exploiting long-range correlation and local information for accurate monocular depth estimation.
\newblock \emph{Machine Intelligence Research}, 2023{\natexlab{b}}.

\bibitem[Lin et~al.(2014)Lin, Maire, Belongie, Hays, Perona, Ramanan, Doll{\'a}r, and Zitnick]{lin2014microsoft}
Tsung-Yi Lin, Michael Maire, Serge Belongie, James Hays, Pietro Perona, Deva Ramanan, Piotr Doll{\'a}r, and C~Lawrence Zitnick.
\newblock Microsoft coco: Common objects in context.
\newblock In \emph{ECCV}, 2014.

\bibitem[Liu et~al.(2021{\natexlab{a}})Liu, Liu, Fan, and Huang]{liu2021polarized}
Huajun Liu, Fuqiang Liu, Xinyi Fan, and Dong Huang.
\newblock Polarized self-attention: Towards high-quality pixel-wise regression.
\newblock \emph{arXiv preprint}, 2021{\natexlab{a}}.

\bibitem[Liu et~al.(2022{\natexlab{a}})Liu, Ren, Lin, and Zhao]{liu2022pseudo}
Luping Liu, Yi Ren, Zhijie Lin, and Zhou Zhao.
\newblock Pseudo numerical methods for diffusion models on manifolds.
\newblock \emph{arXiv preprint}, 2022{\natexlab{a}}.

\bibitem[Liu et~al.(2021{\natexlab{b}})Liu, Lin, Cao, Hu, Wei, Zhang, Lin, and Guo]{liu2021swin}
Ze Liu, Yutong Lin, Yue Cao, Han Hu, Yixuan Wei, Zheng Zhang, Stephen Lin, and Baining Guo.
\newblock Swin transformer: Hierarchical vision transformer using shifted windows.
\newblock In \emph{ICCV}, 2021{\natexlab{b}}.

\bibitem[Liu et~al.(2022{\natexlab{b}})Liu, Mao, Wu, Feichtenhofer, Darrell, and Xie]{liu2022convnet}
Zhuang Liu, Hanzi Mao, Chao-Yuan Wu, Christoph Feichtenhofer, Trevor Darrell, and Saining Xie.
\newblock A convnet for the 2020s.
\newblock In \emph{CVPR}, 2022{\natexlab{b}}.

\bibitem[Loshchilov and Hutter(2017)]{loshchilov2017decoupled}
Ilya Loshchilov and Frank Hutter.
\newblock Decoupled weight decay regularization.
\newblock \emph{arXiv preprint}, 2017.

\bibitem[Lu et~al.(2022)Lu, Zhou, Bao, Chen, Li, and Zhu]{lu2022dpm}
Cheng Lu, Yuhao Zhou, Fan Bao, Jianfei Chen, Chongxuan Li, and Jun Zhu.
\newblock Dpm-solver: A fast ode solver for diffusion probabilistic model sampling in around 10 steps.
\newblock In \emph{NeurIPS}, 2022.

\bibitem[Neuhold et~al.(2017)Neuhold, Ollmann, Rota~Bulo, and Kontschieder]{neuhold2017mapillary}
Gerhard Neuhold, Tobias Ollmann, Samuel Rota~Bulo, and Peter Kontschieder.
\newblock The mapillary vistas dataset for semantic understanding of street scenes.
\newblock In \emph{ICCV}, 2017.

\bibitem[Reddi et~al.(2019)Reddi, Kale, and Kumar]{reddi2019convergence}
Sashank~J Reddi, Satyen Kale, and Sanjiv Kumar.
\newblock On the convergence of adam and beyond.
\newblock \emph{arXiv preprint}, 2019.

\bibitem[Rombach et~al.(2022)Rombach, Blattmann, Lorenz, Esser, and Ommer]{rombach2022high}
Robin Rombach, Andreas Blattmann, Dominik Lorenz, Patrick Esser, and Bj{\"o}rn Ommer.
\newblock High-resolution image synthesis with latent diffusion models.
\newblock In \emph{CVPR}, 2022.

\bibitem[Saxena et~al.(2023)Saxena, Kar, Norouzi, and Fleet]{saxena2023monocular}
Saurabh Saxena, Abhishek Kar, Mohammad Norouzi, and David~J Fleet.
\newblock Monocular depth estimation using diffusion models.
\newblock \emph{arXiv preprint}, 2023.

\bibitem[Schuhmann et~al.(2022)Schuhmann, Beaumont, Vencu, Gordon, Wightman, Cherti, Coombes, Katta, Mullis, Wortsman, et~al.]{schuhmann2022laion}
Christoph Schuhmann, Romain Beaumont, Richard Vencu, Cade Gordon, Ross Wightman, Mehdi Cherti, Theo Coombes, Aarush Katta, Clayton Mullis, Mitchell Wortsman, et~al.
\newblock Laion-5b: An open large-scale dataset for training next generation image-text models.
\newblock In \emph{NeurIPS}, 2022.

\bibitem[Silberman et~al.(2012)Silberman, Hoiem, Kohli, and Fergus]{silberman2012indoor}
Nathan Silberman, Derek Hoiem, Pushmeet Kohli, and Rob Fergus.
\newblock Indoor segmentation and support inference from rgbd images.
\newblock In \emph{ECCV}, 2012.

\bibitem[Sohl-Dickstein et~al.(2015)Sohl-Dickstein, Weiss, Maheswaranathan, and Ganguli]{sohl2015deep}
Jascha Sohl-Dickstein, Eric Weiss, Niru Maheswaranathan, and Surya Ganguli.
\newblock Deep unsupervised learning using nonequilibrium thermodynamics.
\newblock In \emph{ICML}, 2015.

\bibitem[Song et~al.(2020)Song, Meng, and Ermon]{song2020denoising}
Jiaming Song, Chenlin Meng, and Stefano Ermon.
\newblock Denoising diffusion implicit models.
\newblock \emph{arXiv preprint}, 2020.

\bibitem[Sun et~al.(2019)Sun, Xiao, Liu, and Wang]{sun2019deep}
Ke Sun, Bin Xiao, Dong Liu, and Jingdong Wang.
\newblock Deep high-resolution representation learning for human pose estimation.
\newblock In \emph{CVPR}, 2019.

\bibitem[Tao et~al.(2020)Tao, Sapra, and Catanzaro]{tao2020hierarchical}
Andrew Tao, Karan Sapra, and Bryan Catanzaro.
\newblock Hierarchical multi-scale attention for semantic segmentation.
\newblock \emph{arXiv preprint}, 2020.

\bibitem[Wang et~al.(2023)Wang, Dai, Chen, Huang, Li, Zhu, Hu, Lu, Lu, Li, et~al.]{wang2023internimage}
Wenhai Wang, Jifeng Dai, Zhe Chen, Zhenhang Huang, Zhiqi Li, Xizhou Zhu, Xiaowei Hu, Tong Lu, Lewei Lu, Hongsheng Li, et~al.
\newblock Internimage: Exploring large-scale vision foundation models with deformable convolutions.
\newblock In \emph{CVPR}, 2023.

\bibitem[Xiao et~al.(2018{\natexlab{a}})Xiao, Wu, and Wei]{xiao2018simple}
Bin Xiao, Haiping Wu, and Yichen Wei.
\newblock Simple baselines for human pose estimation and tracking.
\newblock In \emph{ECCV}, 2018{\natexlab{a}}.

\bibitem[Xiao et~al.(2018{\natexlab{b}})Xiao, Liu, Zhou, Jiang, and Sun]{xiao2018unified}
Tete Xiao, Yingcheng Liu, Bolei Zhou, Yuning Jiang, and Jian Sun.
\newblock Unified perceptual parsing for scene understanding.
\newblock In \emph{ECCV}, 2018{\natexlab{b}}.

\bibitem[Xie et~al.(2021)Xie, Wang, Yu, Anandkumar, Alvarez, and Luo]{xie2021segformer}
Enze Xie, Wenhai Wang, Zhiding Yu, Anima Anandkumar, Jose~M Alvarez, and Ping Luo.
\newblock Segformer: Simple and efficient design for semantic segmentation with transformers.
\newblock In \emph{NeurIPS}, 2021.

\bibitem[Xie et~al.(2023)Xie, Geng, Hu, Zhang, Hu, and Cao]{xie2023revealing}
Zhenda Xie, Zigang Geng, Jingcheng Hu, Zheng Zhang, Han Hu, and Yue Cao.
\newblock Revealing the dark secrets of masked image modeling.
\newblock In \emph{Proceedings of the IEEE/CVF Conference on Computer Vision and Pattern Recognition}, pages 14475--14485, 2023.

\bibitem[Xu et~al.(2022)Xu, Zhang, Zhang, and Tao]{xu2022vitpose}
Yufei Xu, Jing Zhang, Qiming Zhang, and Dacheng Tao.
\newblock Vitpose: Simple vision transformer baselines for human pose estimation.
\newblock \emph{NeurIPS}, 2022.

\bibitem[Yang et~al.(2023)Yang, Ma, Ji, and Ren]{yang2023gedepth}
Xiaodong Yang, Zhuang Ma, Zhiyu Ji, and Zhe Ren.
\newblock Gedepth: Ground embedding for monocular depth estimation.
\newblock In \emph{ICCV}, 2023.

\bibitem[Yasarla et~al.(2023)Yasarla, Cai, Jeong, Shi, Garrepalli, and Porikli]{yasarla2023mamo}
Rajeev Yasarla, Hong Cai, Jisoo Jeong, Yunxiao Shi, Risheek Garrepalli, and Fatih Porikli.
\newblock Mamo: Leveraging memory and attention for monocular video depth estimation.
\newblock In \emph{ICCV}, 2023.

\bibitem[Yuan et~al.(2020)Yuan, Chen, and Wang]{yuan2020object}
Yuhui Yuan, Xilin Chen, and Jingdong Wang.
\newblock Object-contextual representations for semantic segmentation.
\newblock In \emph{ECCV}, 2020.

\bibitem[Yuan et~al.(2021)Yuan, Fu, Huang, Lin, Zhang, Chen, and Wang]{yuan2021hrformer}
Yuhui Yuan, Rao Fu, Lang Huang, Weihong Lin, Chao Zhang, Xilin Chen, and Jingdong Wang.
\newblock Hrformer: High-resolution transformer for dense prediction.
\newblock In \emph{NeurIPS}, 2021.

\bibitem[Zhao et~al.(2023)Zhao, Rao, Liu, Liu, Zhou, and Lu]{zhao2023unleashing}
Wenliang Zhao, Yongming Rao, Zuyan Liu, Benlin Liu, Jie Zhou, and Jiwen Lu.
\newblock Unleashing text-to-image diffusion models for visual perception.
\newblock In \emph{ICCV}, 2023.

\bibitem[Zheng et~al.(2021)Zheng, Lu, Zhao, Zhu, Luo, Wang, Fu, Feng, Xiang, Torr, et~al.]{zheng2021rethinking}
Sixiao Zheng, Jiachen Lu, Hengshuang Zhao, Xiatian Zhu, Zekun Luo, Yabiao Wang, Yanwei Fu, Jianfeng Feng, Tao Xiang, Philip~HS Torr, et~al.
\newblock Rethinking semantic segmentation from a sequence-to-sequence perspective with transformers.
\newblock In \emph{CVPR}, 2021.

\bibitem[Zhou et~al.(2017)Zhou, Zhao, Puig, Fidler, Barriuso, and Torralba]{zhou2017scene}
Bolei Zhou, Hang Zhao, Xavier Puig, Sanja Fidler, Adela Barriuso, and Antonio Torralba.
\newblock Scene parsing through ade20k dataset.
\newblock In \emph{CVPR}, 2017.

\end{thebibliography}
}

\newpage
{\huge \textbf{Appendix}}
\appendix
\section{Implementation details}
\subsection{Semantic segmentation}
The default setting is in Table~\ref{tab:seg_setting}. Our implementation is based on public codebase \texttt{mmsegmentation}~\cite{contributors2020mmsegmentation}. Following~\cite{zhao2023unleashing}, we employ the common data augmentation strategies, like random cropping and random flipping.
\begin{table}[ht]
\centering
\begin{tabular}{l|l}
\hline

\hline

\hline
\textbf{config}             & \textbf{value} \\ 
\hline
optimizer                   & AdamW~\cite{loshchilov2017decoupled}     \\
learning rate          & 8e-5         \\
weight decay                & 0.001           \\
batch size                  & 16           \\
learning rate schedule      & Poly \\
warmup iters               & 1500        \\
number of meta prompts    & 150 \\
\hline

\hline

\hline
\end{tabular}
\caption{Semantic segmentation setting.}
\label{tab:seg_setting}
\end{table}

\subsection{Depth estimation}
The setting of depth estimation is in Table~\ref{tab:depth_setting}. 
\begin{table}[ht]
\centering
\begin{tabular}{l|l}
\hline

\hline

\hline
\textbf{config}             & \textbf{value} \\ 
\hline
optimizer                   & AdamW~\cite{loshchilov2017decoupled}     \\
learning rate          & 5e-4         \\
optimizer momentum          & $\beta_1, \beta_2=0.9, 0.999$\\
weight decay                & 0.1           \\
batch size                  & 24           \\
number of meta prompts    & 50 \\
\hline

\hline

\hline
\end{tabular}
\caption{Depth estimation setting.}
\label{tab:depth_setting}
\end{table}

\subsection{Pose estimation}
The default setting is in Table~\ref{tab:pose_setting}. Our implementation is based on public codebase \texttt{mmpose}~\cite{mmpose2020}.
\begin{table}[ht]
\centering
\begin{tabular}{l|l}
\hline

\hline

\hline
\textbf{config}             & \textbf{value} \\ 
\hline
optimizer                   & AdamW~\cite{loshchilov2017decoupled}     \\
learning rate          & 5e-4         \\
optimizer momentum          & $\beta_1, \beta_2=0.9, 0.999$\\
weight decay                & 0.1           \\
batch size                  & 64           \\
learning rate schedule      & Step \\
warmup iters               & 500        \\
number of meta prompts    & 150 \\
\hline

\hline

\hline
\end{tabular}
\caption{Pose estimation setting.}
\label{tab:pose_setting}
\end{table}

\section{Visualization}
\begin{figure}[ht]
    \centering
    \begin{subfigure}[b]{0.11\textwidth}
        \includegraphics[width=\textwidth]{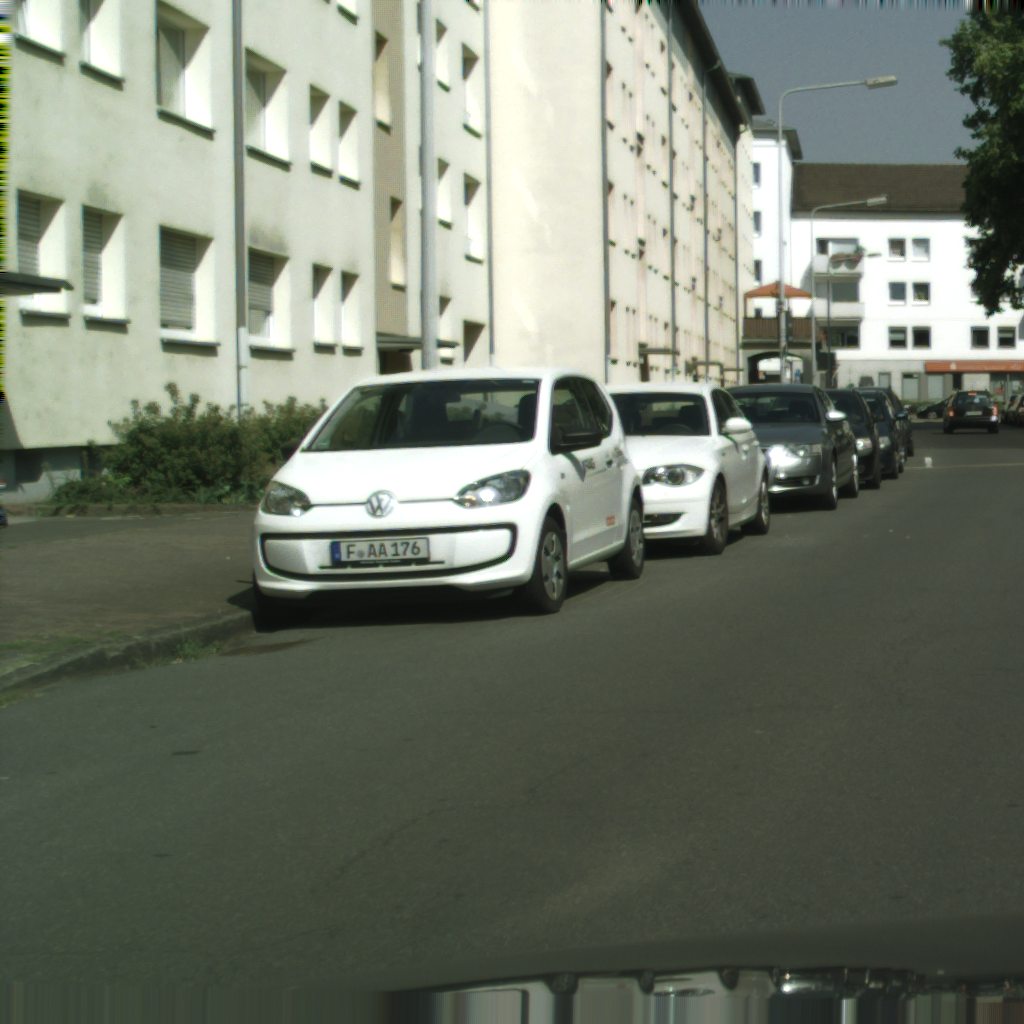}
        \caption{The input image}
    \end{subfigure}
    \begin{subfigure}[b]{0.11\textwidth}
        \includegraphics[width=\textwidth]{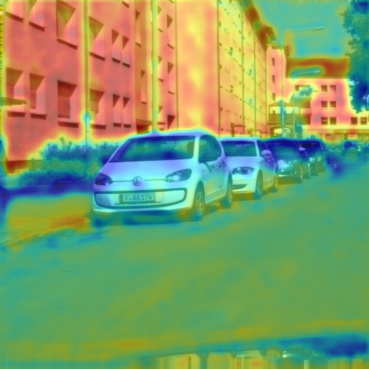}
        \caption{Heatmap of 13th prompt}
    \end{subfigure}
    \begin{subfigure}[b]{0.11\textwidth}
        \includegraphics[width=\textwidth]{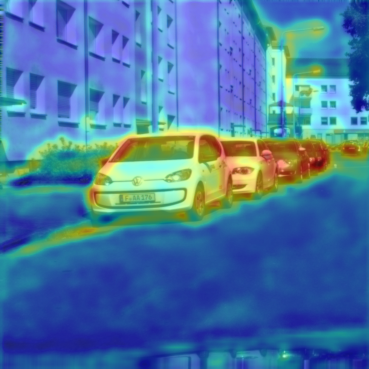}
        \caption{Heatmap of 36th prompt}
    \end{subfigure}
    \begin{subfigure}[b]{0.11\textwidth}
        \includegraphics[width=\textwidth]{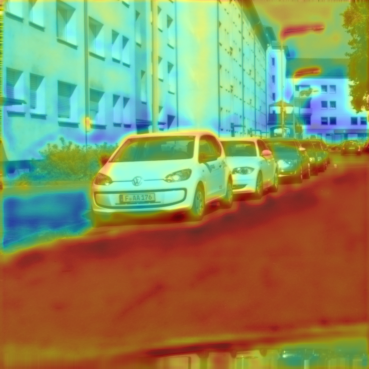}
        \caption{Heatmap of 114th prompt}
    \end{subfigure}

    \begin{subfigure}[b]{0.11\textwidth}
        \includegraphics[width=\textwidth]{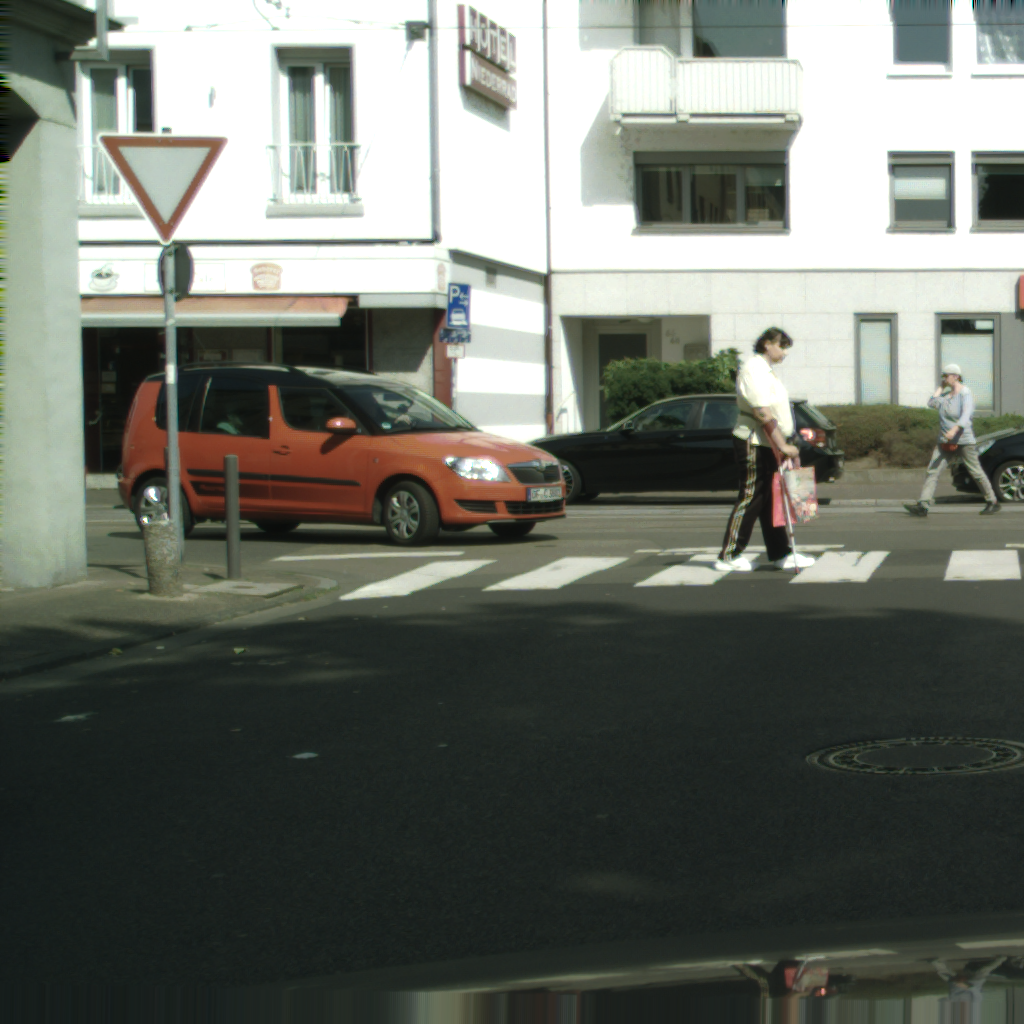}
        \caption{The input image}
    \end{subfigure}
    \begin{subfigure}[b]{0.11\textwidth}
        \includegraphics[width=\textwidth]{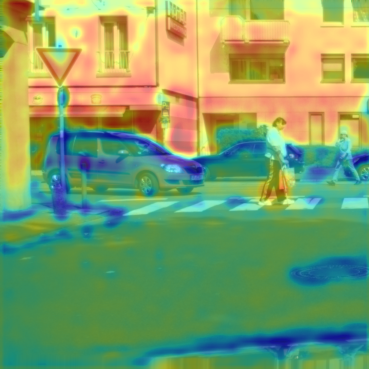}
        \caption{Heatmap of 13th prompt}
    \end{subfigure}
    \begin{subfigure}[b]{0.11\textwidth}
        \includegraphics[width=\textwidth]{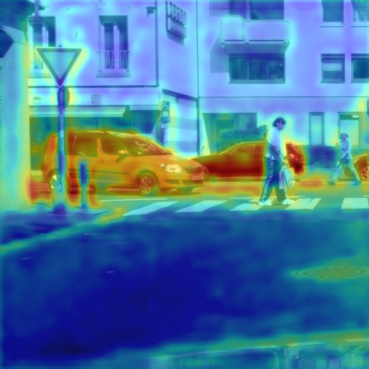}
        \caption{Heatmap of 36th prompt}
    \end{subfigure}
    \begin{subfigure}[b]{0.11\textwidth}
        \includegraphics[width=\textwidth]{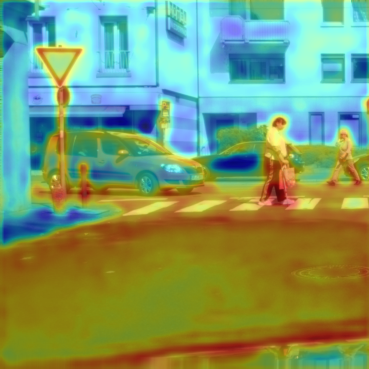}
        \caption{Heatmap of 114th prompt}
    \end{subfigure}
    
    \caption{Heatmaps of meta prompts for semantic segmentation task.}
    \label{fig:sup_heatmap_seg}
\end{figure}

\begin{figure}[ht]
    \centering
    \begin{subfigure}[b]{0.11\textwidth}
        \includegraphics[width=\textwidth]{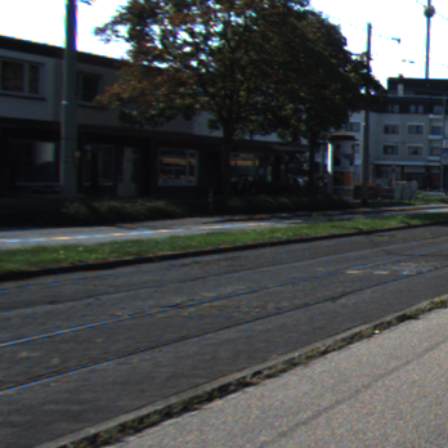}
        \caption{The input image}
    \end{subfigure}
    \begin{subfigure}[b]{0.11\textwidth}
        \includegraphics[width=\textwidth]{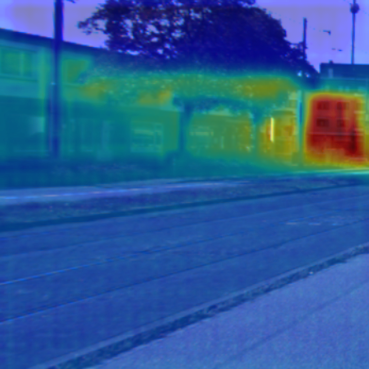}
        \caption{Heatmap of 32th prompt}
    \end{subfigure}
    \begin{subfigure}[b]{0.11\textwidth}
        \includegraphics[width=\textwidth]{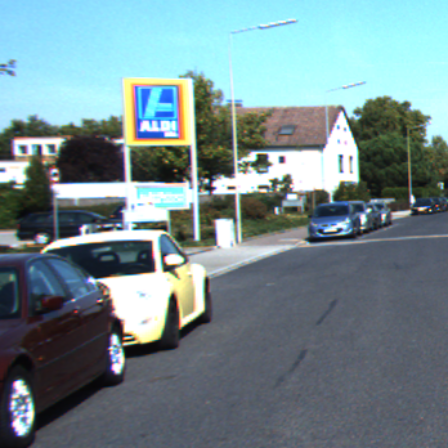}
        \caption{The input image}
    \end{subfigure}
    \begin{subfigure}[b]{0.11\textwidth}
        \includegraphics[width=\textwidth]{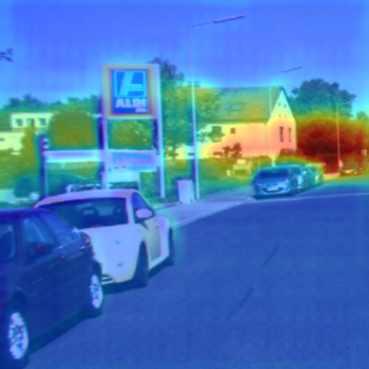}
        \caption{Heatmap of 32th prompt}
    \end{subfigure}

    \begin{subfigure}[b]{0.11\textwidth}
        \includegraphics[width=\textwidth]{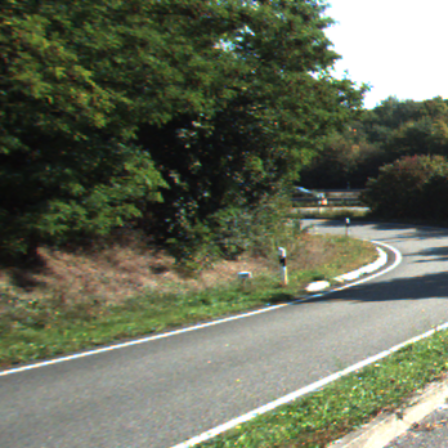}
        \caption{The input image}
    \end{subfigure}
    \begin{subfigure}[b]{0.11\textwidth}
        \includegraphics[width=\textwidth]{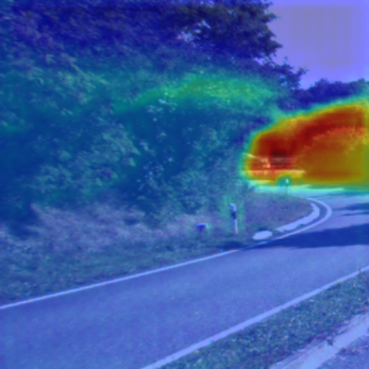}
        \caption{Heatmap of 32th prompt}
    \end{subfigure}
    \begin{subfigure}[b]{0.11\textwidth}
        \includegraphics[width=\textwidth]{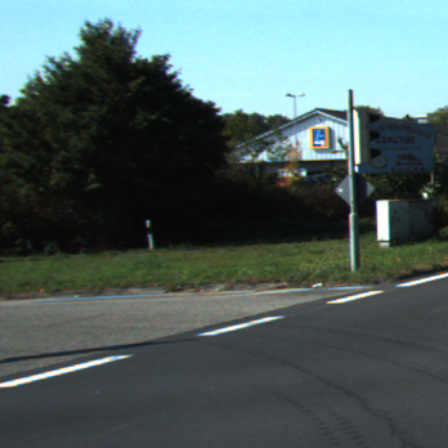}
        \caption{The input image}
    \end{subfigure}
    \begin{subfigure}[b]{0.11\textwidth}
        \includegraphics[width=\textwidth]{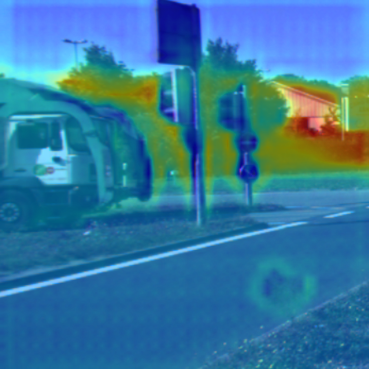}
        \caption{Heatmap of 32th prompt}
    \end{subfigure}
    
    \caption{Heatmaps of meta prompts for depth estimation task.}
    \label{fig:supp_heatmaps_depth}
\end{figure}

\begin{figure}[ht]
    \centering
    \begin{subfigure}[b]{0.11\textwidth}
        \includegraphics[width=\textwidth]{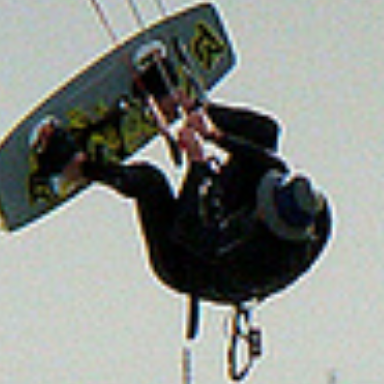}
        \caption{The input image}
        \label{fig:sup_img0}
    \end{subfigure}
    \begin{subfigure}[b]{0.11\textwidth}
        \includegraphics[width=\textwidth]{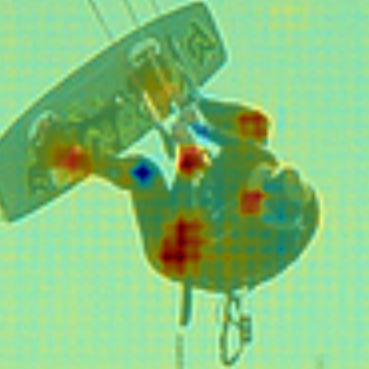}
        \caption{Heatmap of 43th prompt}
        \label{fig:sup_heatmap0}
    \end{subfigure}
    \begin{subfigure}[b]{0.11\textwidth}
        \includegraphics[width=\textwidth]{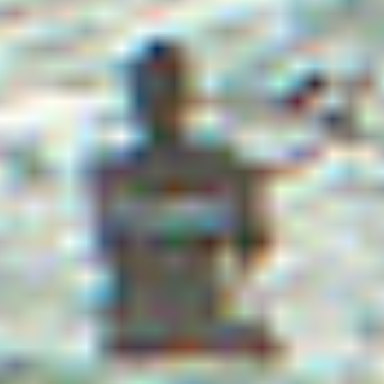}
        \caption{The input image}
        \label{fig:sup_image1}
    \end{subfigure}
    \begin{subfigure}[b]{0.11\textwidth}
        \includegraphics[width=\textwidth]{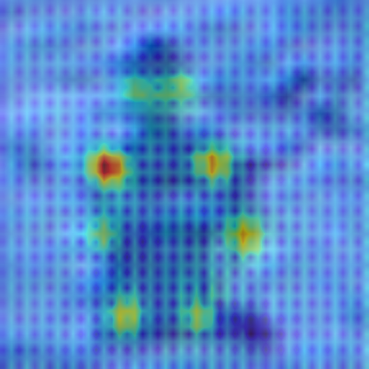}
        \caption{Heatmap of 43th prompt}
        \label{fig:sup_heatmap1}
    \end{subfigure}
    
    \caption{Heatmaps of meta prompts for pose estimation task.}
    \label{fig:sup_heatmap_pose}
\end{figure}

\subsection{Heatmap of meta prompts for pose estimation}
In the main paper, we visualized the heatmaps of meta prompts for semantic segmentation and depth estimation tasks, showcasing their capacity to activate task-relevant features. In this section, we provide additional visualization examples and extend this visualization to pose estimation task. 

As shown in Figure~\ref{fig:sup_heatmap_seg}, for semantic segmentation task, the meta prompts effectively demonstrate class-aware capabilities, essential for pixel-level classification. From the qualitative results, we can find that the same meta prompts tend to activate features of the same category.

For depth estimation task, the meta prompts exhibit depth awareness, with activation values changing based on depth, enabling the prompts to focus on objects at consistent distances, as illustrated in Figure~\ref{fig:supp_heatmaps_depth}. 

As demonstrated in Figure~\ref{fig:sup_heatmap_pose}, the qualitative results show that in pose estimation, meta prompts reveal a different set of capabilities, notably keypoints awareness, which aids in detecting keypoints. These qualitative results collectively highlight the task-relevant activation ability of our proposed meta prompts across various tasks.


\end{document}